\documentclass{Interspeech}

\interspeechcameraready

\title{Alzheimer’s Dementia Detection \\Using Perplexity from Paired Large Language Models}

\author[affiliation={1}]{Yao}{Xiao}
\author[affiliation={1}]{Heidi}{Christensen}
\author[affiliation={1,2}]{Stefan}{Goetze}

\affiliation{School of Computer Science}{The University of Sheffield}{United Kingdom}
\affiliation{}{South Westphalia University of Applied Sciences}{Germany}
\email{yxiao57@sheffield.ac.uk, heidi.christensen@sheffield.ac.uk, s.goetze@sheffield.ac.uk}

\usepackage{comment}

\usepackage{cite}
\usepackage{amsmath,amssymb,amsfonts}
\usepackage{algorithmic}
\usepackage{graphicx}
\usepackage{textcomp}
\usepackage{xcolor}
\usepackage{url}

\usepackage{flushend} 
\usepackage{acronym}  
\usepackage{booktabs}
\usepackage{array}
\acrodef{AD}    {Alzheimer's dementia}
\acrodef{ADR}   {Active Data Representation}
\acrodef{AI}    {artificial intelligence}
\acrodef{AUC}   {Area Under the  \ac{ROC} Curve}
\acrodef{CNN}  {Convolutional Neural Network}
\acrodef{DB}   {DementiaBank}
\acrodef{EER}   {Equal Error Rate}
\acrodef{HC}   {healthy control}
\acrodef{ICL}   {in-context learning}
\acrodef{IPA}   {International Phonetic Alphabet}
\acrodef{LM}    {language model}
\acrodef{LLM}   {large language model}
\acrodef{LoRA}  {Low-Rank Adaptation}
\acrodef{LSTM}   {long short-term memory}
\acrodef{MMSE}   {Mini-Mental State Examination}
\acrodef{POS}    {Part of Speech}
\acrodef{RNN}   {Recurrent Neural Network}
\acrodef{ROC}   {Receiver Operating Characteristic}
\acrodef{SPELL}   {Selecting Prompts by Estimating LM Likelihood}
\acrodef{SVM}   {Support Vector Machine}
\acrodef{SD}   {standard deviation}
\acrodef{TRL}  {Transformer Reinforcement Learning}

\usepackage{multirow, adjustbox}
\usepackage[english]{babel}
\graphicspath{ {./images/} }
\addto\extrasenglish{}
\addto\extrasenglish{}
\addto\extrasenglish{}
\addto\extrasenglish{}
\addto\extrasenglish{}
\addto\extrasenglish{}
\usepackage{graphicx}

\begin{document}

\maketitle

\begin{abstract}
Alzheimer’s dementia (AD) is a neurodegenerative disorder with cognitive decline that commonly impacts language ability. This work extends the paired perplexity approach to detecting AD by using a recent large language model (LLM), the instruction-following version of Mistral-7B. We improve accuracy by an average of $3.33\%$ over the best current paired perplexity method and by $6.35\%$ over the top-ranked method from the ADReSS $2020$ challenge benchmark. Our further analysis demonstrates that the proposed approach can effectively detect AD with a clear and interpretable decision boundary in contrast to other methods that suffer from opaque decision-making processes. Finally, by prompting the fine-tuned LLMs and comparing the model-generated responses to human responses, we illustrate that the LLMs have learned the special language patterns of AD speakers, which opens up possibilities for novel methods of model interpretation and data augmentation.

\end{abstract}
\section{Introduction}
\acresetall
\label{sec:introduction}
Individuals with \ac{AD} often exhibit early signs of declined language ability, particularly in retrieving semantic knowledge. This may result in reduced information content and increased pauses due to word search difficulty, among other indicators \cite{
fraser2016linguistic, Lopez2018
}. Using language to automatically detect \ac{AD} is an appealing alternative to other diagnostic methods, such as neuroimaging and cognitive assessments, due to its non-invasiveness and cost-effectiveness~\cite{Qi2023,AcousticUserInterfaces2010}. 

The \ac{DB} Pitt corpus \cite{becker1994natural} contains responses from individuals with and without \ac{AD} to various cognitive tasks, including the \emph{Cookie Theft} picture description task, where the participant is asked to describe an image of two children stealing cookies. To facilitate research, the ADReSS $2020$ challenge \cite{adress2020} proposed a balanced subset \cite{luz20_interspeech} of \ac{DB} with a standardised train-test split as an \ac{AD} detection benchmark. 

Existing automated \ac{AD} detection methods include using various linguistic and acoustic features, such as Mel-frequency cepstral coefficients (MFCCs) \cite{fraser2016linguistic, Lopez2018,
balagopalan20, shah2021learning, meghanani2021exploration,pan2025two}, vocabulary richness \cite{
fraser2016linguistic, balagopalan20, shah2021learning} or various others, together with classifiers like Random Forest \cite{balagopalan20, shah2021learning, Martinc2021, haulcy2021classifying} or neural network architectures like \ac{LSTM}\cite{mirheidari18_interspeech, meghanani2021exploration, li2021comparative}. With the recent rise of \acp{LLM}, more research has focused on leveraging their capabilities for \ac{AD} detection, primarily through three approaches: (i) using pre-trained \acp{LLM}' embeddings as input features for a downstream classifier \cite{li2021comparative, haulcy2021classifying}, (ii) fine-tuning an \ac{LLM} with a classification layer \cite{balagopalan20, yuan2020, pan21c_interspeech, pappagari21_interspeech, Casu2024}, and (iii) prompting an \ac{LLM} to generate a classification label or analysis based on the participant's transcribed speech, often using zero- or few-shot learning \cite{Balamurali2024, bang2024alzheimer, botelho24macro}. However, methods (i)-(ii) lack interpretability in their decision-making process \cite{botelho24macro,li2022}, and method (iii) is prone to misclassification -- both are crucial concerns in healthcare applications.\looseness=-1

An alternative \ac{AD} detection method is based on \emph{paired perplexity}. Although perplexity is primarily used as an evaluation metric for language models, it has also been shown to reliably characterise individuals' language \cite{colla22}. This capability has been explored for \ac{AD} detection using various models, including n-gram models \cite{wankerl17_interspeech, liu2019dementia, colla22, Chuheng2022}, \acp{LSTM} \cite{klumpp2018, fritsch2019, Chuheng2022}, neural networks with attention mechanism~\cite{Chuheng2022} and GPT-$2$ \cite{colla22, francesco2025}, with a trend of improved classification performance with the use of more sophisticated language models. Specifically, using GPT-$2$ can achieve $100\%$ accuracy for binary classification on \ac{DB}~\cite{colla22} and substantially outperform fine-tuned BERT-based classifiers \cite{francesco2025}. However, limited research has applied the paired perplexity method to models beyond GPT-$2$ \cite{zhu2024adversarial}, despite \acp{LLM} having evolved through several generations since GPT-$2$'s release. 
This work benchmarks and extends the current top-performing \ac{AD} detection methods, proposes a balanced subset of \ac{DB}, fine-tunes the more advanced instruction-following \acp{LLM} for the paired perplexity method, and prompts the fine-tuned \acp{LLM} to inspect the models' learned language patterns as a novel interpretability technique. \looseness=-1



\section{Automated Alzheimer's Dementia Detection with Paired Perplexity}
\label{sec:ad-classification}

Introduced by \cite{wankerl17_interspeech}, the paired perplexity approach to \ac{AD} classification uses a pair of language models, with an \ac{AD} model $M_{\mathrm{AD}}$ trained on transcripts of \ac{AD} speakers, and an \ac{HC} model $M_{\mathrm{C}}$ trained on transcripts of \ac{HC} speakers. 

Perplexity measures how well a language model predicts a sequence. For a given model with a probability distribution $P$ and a sequence $W$ of $N$ words $w_1, \dots, w_N$, the perplexity is defined as
\begin{equation}
  \operatorname{PPL}(W) = P(W)^{-\frac{1}{N}} = \exp\left(-\frac{1}{N} \sum_{i=1}^{N} \log P(w_i)\right).
\end{equation}

For classification, the perplexities of models $M_{\mathrm{AD}}$ and $M_{\mathrm{C}}$ on the test samples, denoted as $\mathrm{PPL_{AD}}$ and $\mathrm{PPL_{C}}$, are computed, and the difference between the two perplexities is used to classify the test sample. The hypothesis is that a test sample coming from an \ac{HC} speaker will be assigned a higher probability by the model trained on the language of \ac{HC} speakers (i.e.\,$M_{\mathrm{C}}$) compared to $M_{\mathrm{AD}}$, which means a lower $\mathrm{PPL_{C}}$ and a higher $\mathrm{PPL_{AD}}$, and vice versa. This method is conceptually straightforward and has a relatively transparent decision-making process.  

\subsection{Paired perplexity for AD detection with LLMs}
\label{ssec:ad_detection_llm}
An impressive accuracy of $100\%$ for the binary AD classification task was reported by \cite{colla22}, using paired perplexity with GPT-$2$ on \ac{DB}. However, \ac{DB} is an unbalanced dataset in terms of age, gender, and class, meaning that the results might be a biased representation \cite{luz20_interspeech}. Furthermore, \cite{colla22} used a subset of \ac{DB} that only included speakers with multiple transcripts, which were collected longitudinally \cite{becker1994natural}. Each participant's class was determined by the perplexity difference averaged across all of the participant's transcripts. Thus, this measure required multiple cognitive assessment responses to be collected over several years, which might not be practical for most \ac{AD} detection applications. It remains unknown whether the approach would be as effective when longitudinal information is not available to the models. Therefore, this work selects \cite{colla22} as one of the baselines and benchmarks it on the ADReSS $2020$ dataset, which is balanced and contains only one transcript per subject, as well as a balanced subset of \ac{DB} created for this study. 

In recent years, \acp{LLM} have evolved into instruction-following models. While earlier models like GPT-$2$ can generate text by continuing a given prompt, they are not specifically tuned for executing instructions in a prompt. However, minimal research has applied instruction-following \acp{LLM} to the paired perplexity method for \ac{AD} detection. This lack of research is addressed in \cite{zhu2024adversarial}, with a focus on \ac{ICL}, which is a task adaptation method that includes task-specific demonstrations in the input prompt. The reported accuracies in \cite{zhu2024adversarial} are $85.42\%$ on ADReSS and $72.27\%$ on \ac{DB}, which do not surpass either of our chosen baselines (see \autoref{sec:baselines}). Therefore, results from \cite{zhu2024adversarial} are not included as a baseline in this work. Furthermore, \ac{ICL} is highly sensitive to the format and number of demonstrations and underperforms fine-tuning on text classification tasks \cite{mosbachetal2023shot}. Fine-tuning, unlike \ac{ICL}, updates model weights and might lead to more substantial change in model perplexity on the adapted tasks. Therefore, this work focuses on fine-tuning \acp{LLM}, rather than using \ac{ICL}, aiming for more accurate \ac{AD} classification. 

Another advantage of introducing fine-tuned instruction-following \acp{LLM} is that they can be prompted to generate synthetic text, which opens up possibilities for novel methods of model interpretation and data augmentation. Specifically, this work probes the language pattern learned by the \acp{LLM} by prompting the fine-tuned models to generate Cookie Theft responses, treating the \acp{LLM} as participants doing the picture description task. This is an attested interpretability technique that has been applied to other areas such as \ac{LLM} reasoning \cite{binz2023using}. 

This work uses the instruction-following version of Mistral-$7$B\footnote{\url{https://huggingface.co/mistralai/Mistral-7B-Instruct-v0.3}} \cite{jiang2023mistral}, an open-source model that has shown superior performance in previous \ac{AD} detection studies \cite{Casu2024, botelho24macro} compared to other \acp{LLM} of its size, even much larger ones.
\subsection{Variants of perplexity differences}

Existing studies have explored various perplexity differences, the most commonly used one being the plain perplexity difference $D = \mathrm{PPL_{AD}} - \mathrm{PPL_{C}}$\cite{wankerl17_interspeech, liu2019dementia, Chuheng2022, klumpp2018, fritsch2019, cohen2020tale}. 
Another study \cite{li2022} has used the perplexity ratio, the logarithmic perplexity difference and its normalised variant. While this study experimented with all these scores, it was found that they produced very similar results to each other. Therefore, only the best-performing one, the normalised logarithmic perplexity difference 
\begin{equation}
  \label{eq:NormLogPerplexityDiff}
\tilde{D}_{\log} = \frac{\log(\mathrm{PPL_{C}}) - \log(\mathrm{PPL_{AD}})}{\log(\mathrm{PPL_{C}})}
\end{equation}
is used in this work in addition to the scores used by the baseline study \cite{colla22} as introduced in the following, ie.~the mean of the plain perplexity differences $D$ for all \ac{AD} speakers  $\overline{D}_{\mathrm{AD}}$ and \ac{HC} speakers $\overline{D}_{\mathrm{C}}$ in the train set. As mentioned in \autoref{ssec:ad_detection_llm}, the data used in \cite{colla22} has multiple transcripts per subject. For a test subject $s$, their average perplexity difference $\overline{D}_{s}$ across all transcripts is calculated. The subject $s$ is classified as having \ac{AD} if $\overline{D}_{\mathrm{AD}}$ is numerically closer to $\overline{D}_{s}$ than $\overline{D}_{\mathrm{C}}$ is, otherwise $s$ is classified as an \ac{HC} speaker.
\begin{equation}
\operatorname{class}(s)=\underset{x \in\{\mathrm{C, AD\}}}{\operatorname{argmin}}\left|\overline{D}_{s}-\overline{D}_x\right|
\end{equation}
For a more lenient threshold, $\overline{D}_x$ can be substituted with $\overline{D}_x^\ast$, which takes the \acp{SD} of $D$ into account:
\begin{equation}
\overline{D}_{\mathrm{C}}^\ast = \overline{D}_{\mathrm{C}} - 2\cdot\sigma_{D_{\mathrm{C}}},~
\overline{D}_{\mathrm{AD}}^\ast = \overline{D}_{\mathrm{AD}} + 2\cdot\sigma_{D_{\mathrm{AD}}}
  \end{equation}
In this work, to mitigate the bias from longitudinal data, the classification is performed per transcript rather than per speaker. Thus, the perplexity difference of each subject's transcript ${D}_{s}$ is used instead of $\overline{D}_{s}$. Additionally, for consistency in evaluation, the fixed cutoffs are translated into scores:
\begin{align}
  \label{eq:MeanPerplexityDistance}
  \overline{D} &= \left| D_{s} - \overline{D}_{\mathrm{C}} \right| - \left| D_{s} - \overline{D}_{\mathrm{AD}} \right| \\
  \label{eq:MeanPerplexityDistanceWithStandardDeviation}
  \overline{D}^\ast &= \left| D_{s} - \overline{D}_{\mathrm{C}}^\ast \right| - \left| D_{s} - \overline{D}_{\mathrm{AD}}^\ast \right|
\end{align}

\vspace{-5pt}

\subsection{Baselines}
\label{sec:baselines}
As discussed in \autoref{ssec:ad_detection_llm}, \cite{colla22} is included as a baseline and is denoted as Baseline~A in the following. Additionally, this work refers to the ADReSS $2020$ challenge for a second baseline: in \cite{Martinc2021}, an ensemble of $50$ Random Forest classifiers was trained, with the final labels determined by majority voting. Solely using character $4$-grams as features, this method achieved the same accuracy as the winning team \cite{yuan2020} of the ADReSS challenge. Although audio input further increased accuracy, this work focuses on text-only \ac{AD} detection and thus compares against the text-only performance from \cite{Martinc2021}. Therefore, their character 4-grams configuration is denoted as Baseline~B in the following.

\section{Methodology}
\label{sec:methodology}
\subsection{ADReSS data and selection of DB data}
\label{sec:data}
The participants' transcripts in the ADReSS dataset \cite{luz20_interspeech} were used. Additionally, to create a larger dataset, a subset of \ac{DB} was curated\footnote{The code for generating this subset and implementing the experiments is available at \url{https://github.com/yaoxiao1999/paired-ppl-ad-detection}.}. Since ADReSS is a subset of DB, the following conditions were imposed to prevent data leakage: there are no test speakers in the train set, and vice versa; all transcripts from ADReSS are removed; there are no ADReSS test speakers in the DB train set, and no DB test speakers in the ADReSS train set. The AD and \ac{HC} groups were optimised to balance in terms of gender, age, and years of education, which all have been shown to affect cognitive test performance \cite{lezak2004neuropsychological}. The data was split into train and test sets in $70/30$ ratio, which were also balanced to each other according to the aforementioned metadata as well as \ac{MMSE}. This \ac{DB} subset is slightly more than double the size of ADReSS, containing $228$ transcripts in the train set and $98$ transcripts in the test set, with an even distribution of \ac{AD} and \ac{HC} classes.   



Text preprocessing included removing all manual linguistics annotations, substituting \ac{IPA} symbols with Latin characters, and removing all special characters except full stops and apostrophes.

To fine-tune Mistral, an instruction dataset was created to simulate interviewer-participant conversations. Each entry consists of a fixed prompt that asks the model to describe the Cookie Theft picture, paired with a transcript from the training dataset as the desired model response. 

\subsection{Experimental setup}
For the proposed method, the Mistral models $M_{\mathrm{AD}}$ and $M_{\mathrm{C}}$ were fine-tuned on the subsets of the instruction dataset containing \ac{AD} and \ac{HC} transcripts from the train set, respectively. For Baseline~A, models  $M_{\mathrm{AD}}$ and $M_{\mathrm{C}}$ were obtained by fine-tuning GPT-$2$ on the plain transcripts of \ac{AD} and \ac{HC} speakers in the train set, respectively. During testing, the perplexity differences in (\ref{eq:NormLogPerplexityDiff}), (\ref{eq:MeanPerplexityDistance}) and (\ref{eq:MeanPerplexityDistanceWithStandardDeviation}) were computed using the perplexities of $M_{\mathrm{AD}}$ and $M_{\mathrm{C}}$ on the transcripts in the test set. The transcripts were then classified as either from an \ac{AD} speaker or an \ac{HC} speaker based on the perplexity differences.

In Baseline~B, $50$ Random Forest classifiers were trained using $50$ different random seeds, and the output labels of the $50$ classifiers were aggregated to produce the final classification. For robust evaluation and a fair comparison between methods, the proposed method and Baseline~A were aligned with Baseline~B such that $50$ pairs of models were trained using the same random seeds as in Baseline~B. The mean, best, and aggregated results will be reported in \autoref{sec:results}.

\subsection{Evaluation metrics}
\label{ssec:EvaluationMetrics}
The evaluation metrics for \ac{AD} classification were chosen as in previous paired perplexity studies \cite{klumpp2018, fritsch2019, wankerl17_interspeech, cohen2020tale, Chuheng2022, li2022}, namely (i)~the \ac{AUC}, which quantifies a model's ability to distinguish between classes, with a higher AUC indicating better performance; (ii)~accuracy at the \ac{EER} point, which is the classification accuracy at the threshold where the false positive (FP) rate equals the false negative (FN) rate; and (iii)~the Pearson correlation with \ac{MMSE} $r_{\mathrm{MMSE}}$ \cite{folstein1983mini}, measuring how well the model's outputs correlate with the participants' cognitive assessment scores.

For the proposed Mistral-based method and Baseline~A, the perplexity differences in (\ref{eq:NormLogPerplexityDiff}), (\ref{eq:MeanPerplexityDistance}) and (\ref{eq:MeanPerplexityDistanceWithStandardDeviation}) were used to compute the evaluation metrics. For Baseline~B, the predicted probabilities for the positive class were used instead of Random Forest's predefined classification threshold since the metrics required continuous scores. For the same reason, the aggregation method was to average the perplexity differences or the positive-class probabilities across $50$ models instead of majority voting as in Baseline~B's original work \cite{Martinc2021}. For reference, Baseline~B was also replicated exactly as the original work on ADReSS. The mean ($\pm $\ac{SD}), best, and aggregated accuracy are $86.25\%$ ($\pm 3.18\%$), $93.75\%$, and $91.67\%$, respectively. The decline in accuracy of its modified implementation used in this work (see \autoref{table:acc_auc_r}) is mainly due to the difference in text preprocessing. The original implementation retains the manual linguistic annotations, which provide an advantage to character $4$-grams but are impractical for automated \ac{AD} detection applications.\looseness=-1

\subsection{Implementation details}
For fine-tuning Mistral, the data was partitioned into training and validation sets with a $70/30$ ratio. The default batch size of $8$ was used for training and validation, and validation took place after every training step. The models were fine-tuned for $6$ epochs, with an early-stopping callback with patience of $3$ validations to avoid overfitting. The AdamW \cite{loshchilov2018decoupled} optimiser was used with an initial learning rate of $5.0$e$-05$. \ac{LoRA}~\cite{hu2022lora} was used with a rank of $16$ and $\alpha$ of $32$.  At the end of the training, the \ac{LoRA} adapter with the lowest validation loss was merged with the base model. 

The baselines were implemented based on their provided code\footnote{Baseline\,A:\,\url{https://github.com/davidecolla/semantic_coherence_markers},\,\,\,Baseline\,B:\,\,\url{https://github.com/matejMartinc/alzheimer_diagnosis} }. GPT-$2$ was fine-tuned for $30$ epochs on \ac{DB} as suggested by Baseline~A \cite{colla22}. For ADReSS the epoch number was limited to $5$, because preliminary experiments showed a decline in all evaluation metrics after $5$ epochs. This was likely due to overfitting to ADReSS, a smaller dataset than \ac{DB}. 
\section{Results and analysis}
\label{sec:results}
\subsection{Classification performance}
As \autoref{table:acc_auc_r} shows, in both datasets, the proposed method outperforms both baselines across all evaluation metrics except best and aggregated accuracy on ADReSS and mean \ac{AUC} on \ac{DB}. 

The paired-perplexity-based proposed method and Baseline A perform consistently on both datasets, while Baseline B does not generalise well on \ac{DB}. Interestingly, Baseline~A no longer achieves $100\%$ accuracy on \ac{DB} as in the original work \cite{colla22}, which is likely due to the removal of the advantage from longitudinal data. Among the different perplexity differences, $\tilde{D}_{\log}$ consistently shows better results in terms of all metrics compared to $\overline{D}$ and $\overline{D}^\ast$ for the GPT-2-based Baseline A. While for the proposed Mistral-based method, the performance is more independent of the type of perplexity difference used.

\begin{table*}[ht]
  \caption{Accuracy, \ac{AUC}, and $r$ with \ac{MMSE} of different models and perplexity differences. Best results per model are underlined, and the overall best results are highlighted in boldface. \emph{Agg} denotes aggregated results.}
    \vspace{-6pt}
  \begin{minipage}{\textwidth}
  \begin{adjustbox}{width={\textwidth}, keepaspectratio}%
    {\fontsize{8}{8}\selectfont
  \begin{tabular}{cclllll|llll|rrrr}
      \toprule
      \multirow{2}{*}{Data}
      & \multirow{2}{*}{Model} 
      & \multirow{2}{*}{PPL} 
      & \multicolumn{4}{c|}{Accuracy at EER (\%)} & \multicolumn{4}{c|}{AUC (\%)} & \multicolumn{4}{c}{Pearson's $r$ with MMSE} \\
       & & & Mean & SD & Best & Agg & Mean & SD & Best & Agg & Mean & SD & Best & Agg \\

      \midrule
    \multirow{7}{*}{\rotatebox[origin=c]{90}{ADReSS\hspace{7mm}}}
      &\multirow{3}{*}{\shortstack{GPT-2 \\ (Baseline~A \cite{colla22})}
      } 

      & $\tilde{D}_{\log}$ 
      & $\underline{86.21}$ & $2.73$ & $\underline{89.58}$ & $\underline{87.50}$ 
      & $\underline{91.46}$ & $0.91$ & $\underline{93.40}$ & $\underline{91.84}$ 
      & $\underline{-0.71}$ & $0.01$ & $\underline{-0.74}$ & $\underline{-0.72}$ \\

      & & $\overline{D}$ 
      & $79.88$ & $5.31$ & $87.50$ & $79.17$ 
      & $88.58$ & $3.35$ & $93.23$ & $90.10$ 
      & $0.59$ & $0.05$ & $0.67$ & $0.61$ \\
     
      & & $\overline{D}^\ast$ & 
      $79.63$ & $5.66$ & $87.50$ & $79.17$ & $88.46$ & $3.24$ & $93.23$ & $90.10$ 
      & $0.59$ & $0.05$ & $0.67$ & $0.61$ \\

    \cmidrule[0.05pt](lr){2-15}
      & Baseline~B \cite{Martinc2021} & 
      $-$ & 
      $82.83$ & $4.05$ & $\textbf{95.83}$ & $87.50$ 
      & \textbf{$91.28$} & $1.92$ & $96.35$ & $94.79$ 
      & $-0.73$ & $0.03$ & $-0.79$ & $-0.77$ \\

    \cmidrule[0.05pt](lr){2-15}
      & \multirow{3}{*}{Mistral (proposed)}
      & $\tilde{D}_{\log}$ 
      & $\textbf{86.37}$ & $3.61$ & $93.75$ & $\textbf{91.67}$ & $92.71$ & $1.60$ & $95.31$ & $92.71$ & $-0.69$ & $0.02$ & $-0.74$ & $-0.71$ \\

      & & $\overline{D}$ & $85.96$ & $3.33$ & $93.75$ & $89.58$ & $\textbf{93.91}$ & $1.20$ & $\textbf{97.05}$ & $\textbf{94.97}$ & $\textbf{0.76}$ & $0.02$ & $\textbf{0.82}$ & $\textbf{0.78}$  \\
     
      & & $\overline{D}^\ast$ & $85.96$ & $3.33$ & $93.75$ &
      $87.50$ & $93.65$ & $1.46$ & $\textbf{97.05}$ & $89.76$ & $\textbf{0.76}$ & $0.02$ & $\textbf{0.82}$ & $\textbf{0.78}$  \\
        \midrule[0.5pt]

      \multirow{7}{*}{\rotatebox[origin=c]{90}{DB subset\hspace{7mm}}}
      
       & \multirow{3}{*}{\shortstack{GPT-2 \\ (Baseline~A \cite{colla22})}
      } 
    & $\tilde{D}_{\log}$ 
      & $\underline{84.98}$ & $1.53$ & $\underline{87.76}$ & $\textbf{85.71}$ 
      & $\textbf{90.56}$ & $0.29$ & $\underline{91.13}$ & $\underline{90.71}$ 
      & $\underline{-0.66}$ & $0.04$ & $\underline{-0.67}$ & $\underline{-0.67}$ \\

      & & $\overline{D}$ 
      & $80.30$ & $1.99$ & $84.69$ & $81.63$ 
      & $84.45$ & $0.42$ & $85.38$ & $84.21$ 
      & $0.60$ & $0.01$ & $0.62$ & $0.60$  \\
     
      & & $\overline{D}^\ast$ 
      & $80.18$ & $2.03$ & $84.69$ & $81.63$ 
      & $83.94$ & $0.54$ & $84.96$ & $83.74$ 
      & $0.60$ & $0.01$ & $0.62$ & $0.60$  \\

    \cmidrule[0.05pt](lr){2-15}
        
    & Baseline~B \cite{Martinc2021} &
      $-$ & 
      $75.80$ & $2.94$ & $84.69$ & $77.55$ 
      & $84.00$ & $1.63$ & $87.92$ & $85.55$ 
      & $-0.59$ & $0.03$ & $-0.65$ & $-0.62$ \\

    \cmidrule[0.05pt](lr){2-15}

      & \multirow{3}{*}{Mistral (proposed)}
      & $\tilde{D}_{\log}$ 
      & $84.59$ & $2.12$ & $89.80$ & $82.65$ 
      & $\underline{90.42}$ & $1.30$ & $\textbf{93.29}$ & $\textbf{91.42}$ 
      & $-0.68$ & $0.03$ & $-0.74$ & $-0.71$ \\

      & & $\overline{D}$ 
      & $82.36$ & $2.95$ & $88.78$ & $\underline{83.67}$ 
      & $88.21$ & $1.74$ & $92.04$ & $89.25$       
      & $\textbf{0.70}$ & $0.03$ & $\textbf{0.75}$ & $\textbf{0.72}$  \\
     
      & & $\overline{D}^\ast$
      & $\textbf{85.96}$ & $3.33$ & $\textbf{93.75}$ & $75.51$ 
      & $87.92$ & $1.81$ & $91.77$ & $76.76$ 
      & $\textbf{0.70}$ & $0.03$ & $\textbf{0.75}$ & $\textbf{0.72}$  \\
      
      \bottomrule
      \end{tabular}
      }
  \end{adjustbox}
  \end{minipage}
  \label{table:acc_auc_r}
  \vspace{-5pt}
  \end{table*}

\subsection{Analysing perplexity differences relative to MMSE}

\begin{figure}[ht!]  
   \includegraphics[width=\columnwidth]{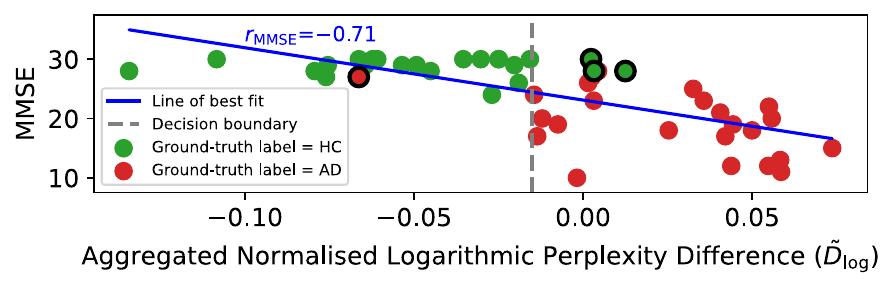}
  \vspace{-20pt}
  \caption{MMSE vs.\ aggregated $\tilde{D}_{\log}$, using ADReSS data. The four missclassified data points are circled in black.}
    \vspace{-15pt}
  \label{fig:scatterplot}
\end{figure}

To illustrate how perplexity differences distinguish between classes, \autoref{fig:scatterplot} shows the distribution of Mistral's aggregated $\tilde{D}_{\log}$ on the ADReSS test set. The line of best fit shows a strong correlation ($r_{\mathrm{MMSE}}$=$-0.71$) between $\tilde{D}_{\log}$ and \ac{MMSE}. A clear decision boundary can be seen. As expected, for most \ac{HC} samples, $\tilde{D}_{\log}$ have low values, indicating that $\mathrm{PPL_{C}} < \mathrm{PPL_{AD}}$, and vice versa. Two out of the three misclassified \ac{HC} participants had \ac{MMSE} scores that were below the average for the \ac{HC} group, which means that their cognitive function may have been closer to participants with \ac{AD}. Similarly, the misclassified \ac{AD} participant had an \ac{MMSE} score that was higher than the average \ac{AD} speakers. This discrepancy between class label and cognitive performance might account for the misclassification.

\subsection{Analysing learned linguistic properties}
Further experiments inspected the linguistic patterns learned by the \acp{LLM}. Among the $50$ pairs of Mistral models fine-tuned for \ac{AD} classification, text was generated by the $3$ best pairs of models according to classification accuracy. Using the prompts in the instruction dataset (see \autoref{sec:data}), each model $M_{\mathrm{AD}}$ generated Cookie Theft responses to simulate the language of \ac{AD} speakers and the same was done for each $M_{\mathrm{C}}$ to simulate \ac{HC} speakers. The number of generated samples matched the size of the dataset on which the models were trained (ADReSS or \ac{DB}).\looseness=-1

\begin{figure}[ht]
\includegraphics[width=\columnwidth]{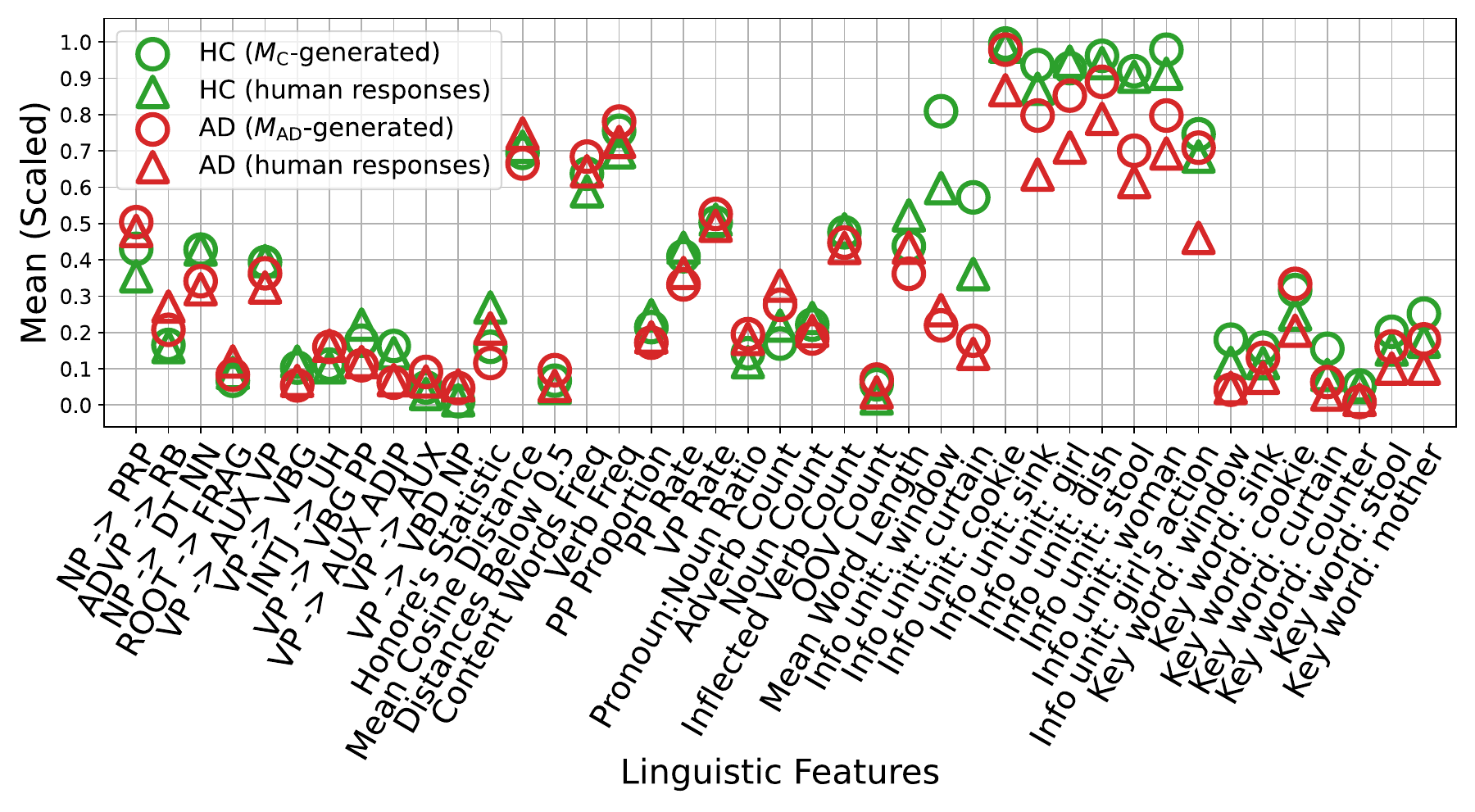}
  \vspace{-20pt}
  \caption{Mean values of linguistic features~\cite{fraser2016linguistic} extracted from Mistral-generated responses and human responses.}
  \vspace{-15pt}
\label{fig:meanfeaturevalues}
\end{figure}

In \cite{fraser2016linguistic}, the top features for distinguishing \ac{AD} speech are identified. This work adopts these features apart from the acoustic features, as this work is text-based only, resulting in $41$ linguistic features. Each feature was extracted from the generated and human-produced (i.e.~ADReSS and DB) responses. Tested by Kolmogorov-Smirnov test, 
the distributions of $30$ out of $41$ features are significantly different between the generated \ac{AD} text and \ac{HC} text,
indicating that the differences in language patterns between \ac{AD} and \ac{HC} speakers have been learned by the \acp{LLM}. \autoref{fig:meanfeaturevalues} shows that for most features, the generated text has a similar mean as the human responses of the same class. \autoref{table:synth_text_example} presents two examples of the generated text. The $M_{\mathrm{AD}}$-generated text reflects \ac{AD} speakers' tendency to have more disfluencies due to word-search difficulty. While the $M_{\mathrm{C}}$-generated text uses full and grammatical sentences and more accurate word choices, as expected for an \ac{HC} speaker. 

The evidence that language patterns are well learned suggests that $M_{\mathrm{AD}}$ and $M_{\mathrm{C}}$ effectively represent the language of their respective classes, leading to low perplexity for sequences that closely match their learned patterns, fulfilling the theoretical basis of the paired perplexity method. This explains why the proposed approach performs effectively in classification. 

\begin{table}[!ht]
\caption{Excerpts of text generated by fine-tuned \acp{LLM}. Signs of cognitive decline that the model has learned are underlined.}
  \vspace{-6pt}
\begin{adjustbox}{width={\columnwidth}}
  \begin{tabular}{@{}p{0.1\columnwidth}p{0.9\columnwidth}}
  \toprule
Model & Generated Text \\
\midrule
$M_{\mathrm{C}}$ & Well let's see. Uh a boy is taking cookies out of the cookie jar. A girl is standing by. The stool is falling over. Mother's drying the dishes. Water's running over. {[}...{]} \\
$M_{\mathrm{AD}}$ & Um. The little boy is on a \underline{wobb. Wobbly} stool getting cookies and the little girl's holding out her hand. And \underline{the lady's. The mother's} down below. {[}...{]}\\
\bottomrule
\end{tabular}
\end{adjustbox}
\label{table:synth_text_example}
  \vspace{-10pt}
\end{table}

\section{Conclusion}
\label{sec:conclusion}
This work extends the paired perplexity approach to automated \ac{AD} detection by incorporating the recent \acp{LLM}, making four key contributions. First, it applies paired perplexity to the newer instruction-following \acp{LLM}, which is an accurate method of \ac{AD} detection with a best accuracy of $93.75\%$, outperforming both the best-performing paired perplexity method \cite{colla22} by an average of $3.33\%$ in accuracy and the top-ranked method \cite{Martinc2021} from the ADReSS $2020$ challenge by $6.35\%$. While many other methods for \ac{AD} detection suffer from opaque decision-making processes, as also pointed out by \cite{botelho24macro,li2022}, the proposed approach is shown to effectively detect \ac{AD} with a clear and interpretable decision boundary, achieving a balance between performance and interpretability. Second, the work addresses the practicality concerns regarding \cite{colla22}'s data selection and \cite{Martinc2021}'s text preprocessing and benchmarks their methods on balanced datasets with the same preprocessing. Third, to promote unbiased and reproducible model training and evaluation, this work proposes a balanced subset of \ac{DB} with a predefined train-test split. Lastly, by prompting the fine-tuned \acp{LLM} to generate Cookie Theft responses, this work shows that the \acp{LLM} have learned the special language patterns of \ac{AD} speakers, which has value for model interpretation and data augmentation.




\bibliographystyle{IEEEtran}
\bibliography{mybib}

\end{document}